\documentclass[12pt]{article}
\usepackage{graphicx}
\usepackage{amsmath}
\usepackage{lipsum}
\usepackage{hyperref}

\title{\textbf{Switch-Based Multi-Part Neural Network}}
\author{
 Surajit Majumder \\
  IBM  \\
  \texttt{surajit.majumder@ibm.com} \\
  \and
 Paritosh Ranjan \\
  IBM  \\
  \texttt{paranjan@in.ibm.com} \\
  \and
 Prodip Roy \\
  IBM  \\
  \texttt{prodipro@in.ibm.com} \\
  \and
 Bhuban Padhan \\
  IBM  \\
  \texttt{  bhubanpadhan@in.ibm.com} \\
}
\date{\today}

\begin{document}

\maketitle

\begin{abstract}
This paper introduces decentralized and modular neural network framework designed to enhance the scalability, interpretability, and performance of artificial intelligence (AI) systems. At the heart of this framework is a dynamic switch mechanism that governs the selective activation and training of individual neurons based on input characteristics, allowing neurons to specialize in distinct segments of the data domain. This approach enables neurons to learn from disjoint subsets of data, mimicking biological brain function by promoting task specialization and improving the interpretability of neural network behavior. Furthermore, the paper explores the application of federated learning and decentralized training for real-world AI deployments, particularly in edge computing and distributed environments. By simulating localized training on non-overlapping data subsets, we demonstrate how modular networks can be efficiently trained and evaluated. The proposed framework also addresses scalability, enabling AI systems to handle large datasets and distributed processing while preserving model transparency and interpretability. Finally, we discuss the potential of this approach in advancing the design of scalable, privacy-preserving, and efficient AI systems for diverse applications.
\end{abstract}

\section{Introduction}

The growing complexity of real-world AI applications, especially those deployed in decentralized or resource-constrained environments such as edge computing, federated learning, and neuro-symbolic systems, has exposed limitations in traditional centralized deep learning paradigms. Conventional models rely on training entire networks on uniformly distributed datasets, assuming uniform data availability and centralized compute capacity. However, such assumptions often fail in practical scenarios where data is fragmented, sensitive to privacy, or originates from diverse, distributed sources.

This research introduces a switch-based multipart neural network, a novel training framework in which individual neurons, or modular groups of neurons, are independently trained on disjoint subsets of data, guided by a dynamic switching mechanism. This switch governs the selective activation and training of specific neurons based on input characteristics or semantic grouping, enabling fine-grained domain-specific learning.

Inspired by the localized learning and task specialization of the brain, this approach goes beyond sparsity and modular selection models by explicitly assigning roles and datasets to neurons, offering unprecedented control and interpretability in neural network training. Each neuron acts as a micro model, learning from partial data exposure while participating in a collaborative evaluation process, resulting in a modular, interpretable, and scalable architecture.

This paradigm has significant implications for scalable AI systems, privacy-preserving learning, and interpretable AI design. It aligns with the increasing need for distributed adaptive learning systems, especially within smart environments such as cities, industrial IoT, or personalized health monitoring, where data locality, heterogeneity, and autonomy are fundamental.

Our implementation in Python using PyTorch demonstrates the feasibility and advantages of this method, such as improved modularity, faster training cycles, and enhanced transparency through per-neuron activation analysis. As a forward-looking innovation, the switch-based architecture holds strong potential for integration into platforms like IBM WatsonX, supporting edge deployments, neuromorphic computing, and hybrid cloud AI solutions.

\section{Brief Description of the Invention}

The core innovation introduced the concept and practical implementation of localized neuron or group of neurons training and execution, where each neuron in a network is trained using distinct, non-overlapping subsets of data and then evaluated collectively. This methodology challenges the traditional assumptions of uniform, global training in neural networks and introduces several novel elements as follows:

\subsection{A. A Dynamic Switching Mechanism}
A switch mechanism is employed during the training process at the neuron level for a single or a group of neurons, wherein specific neurons are selectively activated and updated based on the characteristics of the input data or group of similar inputs as per their configured relationship with the switch. When a neuron participates in processing a particular input subset with an active switch, its associated weights and activation values are updated accordingly. Conversely, neurons that are not engaged during a given training iteration exhibit no activation with an inactive switch. This selective engagement mechanism leads to a natural brain-like assignment of neurons or a group of neurons to specific groups of similar inputs, promoting task specialization and interpretability within the network.

\subsection{B. Localized Intelligence at the Neuron Level}
Rather than treating the neural network as a homogenous system, this approach assigns specific roles or specializations to individual neurons. Each neuron becomes a knowledge unit trained on unique data slices, opening possibilities for:
\begin{itemize}
    \item Context-aware learning
    \item Domain-specific micro-models
\end{itemize}

\subsection{C. Brain-Inspired, Energy-Efficient Computing}
Mirroring biological brains, where neurons learn from limited exposure and collaborate to achieve cognition.

\section{Reduction to Practice}
\begin{figure}[h!]
    \centering
    \includegraphics[width=0.8\textwidth]{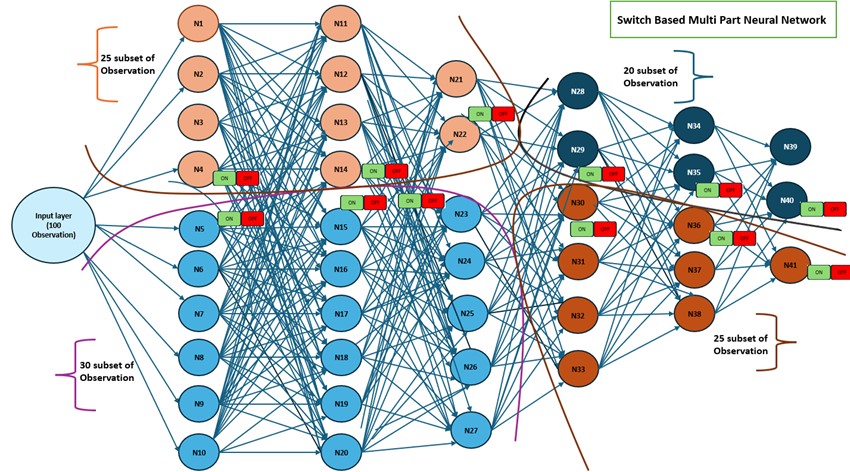}
    \caption{Switch-Based Network Architecture}
    \label{fig:switch_network}
\end{figure}

\begin{figure}[h!]
    \centering
    \includegraphics[width=0.8\textwidth]{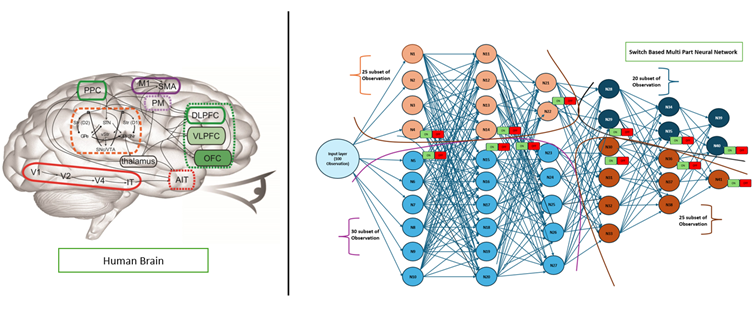}
    \caption{Brain-Inspired Neural Network Architecture}
    \label{fig:brain_vs_network}
\end{figure}
This research introduces a set of innovative methods that collectively redefine how artificial neural networks can be trained, structured, and analysed. Unlike conventional end-to-end training approaches that use uniform backpropagation across all layers and data, the proposed methodology adopts individual neuron or group of neurons-level training with segmented data exposure—resulting in a new class of modular, interpretable, and decentralized neural systems. Below are the key methodological innovations:

\subsection{A. Neuron-Level Data Segmentation and Assignment}
A novel method is applied to split the dataset into non-overlapping subsets that are manually and purposefully assigned to specific neurons in the network. This strategy:
\begin{itemize}
    \item Allows each neuron to specialize in a distinct subset of the data domain.
    \item Introduces functional diversity within the neural network.
    \item Simulates localized learning, inspired by how biological neurons function independently before integrating their outputs.
\end{itemize}

\subsection{B. Independent Localized Training}
Each neuron undergoes isolated training, independent of the rest of the network. This method eliminates cross-neuron gradient sharing, effectively treating each neuron as a standalone learning agent. This is innovative because:
\begin{itemize}
    \item It breaks the traditional paradigm of fully connected and globally optimized layers.
    \item It enables training in parallel and distributed environments, including edge devices or federated nodes.
    \item A dynamic switching mechanism is employed during the training process, wherein specific neurons are selectively activated and updated based on the characteristics of the input data or group of similar inputs. 
\end{itemize}

\subsection{C. Granular Explainability Through Per-Neuron Analysis}
The model supports fine-grained interpretability by observing the behaviour of each neuron on a shared evaluation dataset. Innovative tools and methods are applied, including:
\begin{itemize}
    \item Neuron activation heatmaps
    \item Data group attribution per neuron or group of neurons
    \item Accuracy contribution analysis per neuron or group of neurons
\end{itemize}

This enhances model transparency, which is essential for regulated industries and mission-critical applications.

\section{Process Flow}
The proposed approach of training individual neurons on distinct subsets of data has been implemented through a structured experimental framework that demonstrates its feasibility and potential for broader application in real-world systems. Below is the step-by-step description of the process flow:
\begin{figure}[h!]
    \centering
    \includegraphics[width=0.8\textwidth]{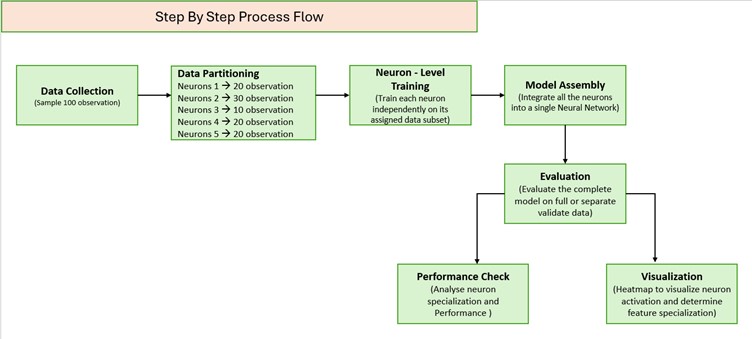}
    \caption{Process Flow of the Training Mechanism}
    \label{fig:process_flow}
\end{figure}
\subsection{A. Dataset Partitioning and Assignment}
A synthetic or real-world dataset containing 100 observations is partitioned into five disjoint subsets, assigned as follows:
\begin{itemize}
    \item Neuron 1: Trained on 20 observations
    \item Neuron 2: Trained on 30 observations
    \item Neuron 3: Trained on 10 observations
    \item Neuron 4: Trained on 20 observations
    \item Neuron 5: Trained on 20 observations
\end{itemize}
Each subset is selected to ensure that it captures a unique, non-overlapping portion of the data distribution, allowing each neuron to specialize in a distinct segment of the problem space. This research investigates a decentralized training approach where each neuron is exposed to a different segment of the dataset.

\subsection{B. Individual Neuron Training}
Each neuron is modeled as a minimalistic unit (e.g., a single-node perceptron or a sub-component of a larger layer) and trained independently using standard gradient descent on its assigned data subset.This simulates a decentralized learning environment where each neuron is trained in isolation—mimicking localized intelligence, like neurons in biological systems or edge devices in a distributed network.

\subsection{C. Centralized Evaluation Mechanism}
Once the training is complete, all neurons are integrated into a single neural network system and evaluated using unseen data, overlapping and non-overlapping test sets, and aggregated performance metrics.

\subsection{D. Analysis and Visualization}
Detailed analysis is conducted using heatmaps to evaluate the performance of each neuron. Activation analysis shows the heatmap visualization provides a clear representation of the activation intensity across neurons for various input categories. Through this analysis, it becomes possible to identify which neurons exhibit higher or lower activation levels in response to specific input types or clusters of similar data. This enables the determination of neuronal responsiveness and specialization, revealing which neurons are most effectively engaged by data groups, thereby offering insights into the model’s internal representation and feature learning behaviour. For instance, Neuron 4 exhibits a higher activation response to the demographic group characterized as Mid-age – Mild Income (Mixed) when compared to Neuron 0, Neuron 1, and Neuron. This indicates that inputs belonging to this specific group are more likely to be selectively processed or represented by Neuron 4. Consequently, Neuron 4 appears to have specialized in recognizing patterns or features associated with this subgroup, suggesting a form of functional differentiation among the neurons based on their training data and resulting activation behaviour.
\begin{figure}[h!]
    \centering
    \includegraphics[width=0.8\textwidth]{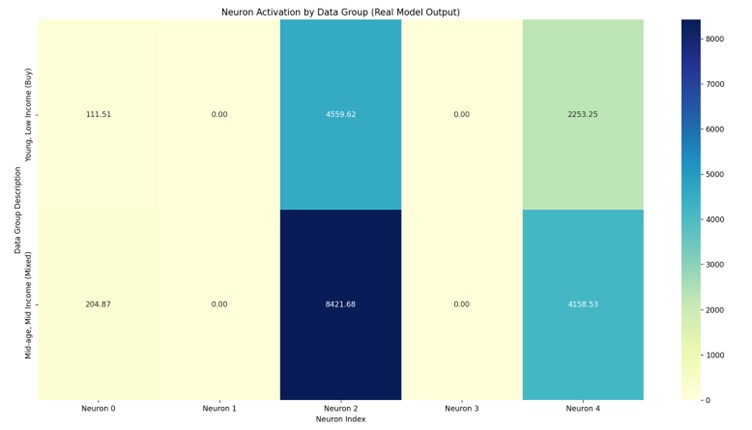}
    \caption{Neuron Activation Heatmap}
    \label{fig:activation_heatmap}
\end{figure}
\subsection{E. Scalability Consideration}
To demonstrate scalability, additional tests can be conducted with larger datasets, increased number of neurons, and deployment in simulated edge environments or federated learning scenarios.

\section{Advantages of the Invention}

The proposed invention offers several advantages over conventional neural network training methods. These advantages span performance, interpretability, and practical deployment:

\subsection{A. Enhanced Specialization and Interpretability}
Each neuron is trained on a unique subset of data, leading to distinct pattern recognition capabilities. This enables:
\begin{itemize}
    \item Clear interpretation of which neuron responds to which data group.
    \item Traceability of learned features to specific data distributions.
    \item Simplified analysis using tools like heatmaps to visualize neuron activations.
\end{itemize}

\subsection{B. Modularity and Scalability}
By decoupling the training process across neurons, the model architecture becomes modular, supporting parallel and distributed training.

\subsection{C. Faster Training Cycles}
Independent training of neurons allows for parallelization across cores or machines and smaller training jobs per neuron, reducing computational load and wall-clock time.

\section{Conclusion}
In this paper, we introduced a framework for decentralized, modular neural networks that leverages a dynamic switching mechanism for neuron specialization. By allowing individual neurons or modular groups of neurons to be trained on distinct subsets of data, the proposed approach addresses several critical challenges in contemporary AI systems, including scalability, interpretability, and task specialization.

Our method simulates localized, brain-inspired learning, where each neuron becomes a micro-model specialized in a particular domain, thereby enhancing the overall performance of the neural network. This enables efficient, parallel training of modular neural networks in distributed environments, making it particularly suitable for decentralized systems such as edge computing and federated learning. Additionally, the approach enhances model transparency by providing granular interpretability at the neuron level, allowing for a deeper understanding of the decision-making process within the network.

Through experimental results, we demonstrated that the switch-based framework not only leads to faster training cycles but also improves the interpretability and scalability of AI models. The ability to partition training data and engage neurons selectively based on input characteristics enables more efficient use of computational resources while maintaining the model's accuracy and robustness.

The proposed architecture holds significant promise for future AI applications, particularly in environments where data privacy, distributed processing, and task-specific specialization are paramount. As AI systems continue to evolve, the switch-based multi-part neural network framework offers a viable path towards creating more modular, interpretable, and scalable models that align with real-world deployment needs.

\section{Acknowledgment}

We would like to express our sincere gratitude to all individuals and organizations who have contributed to the success of this research. We acknowledge the invaluable support from the IBM team, whose resources and expertise have greatly enhanced this project.
Special thanks to Prodip Roy (Program Manager IBM) for their insightful feedback, guidance, and encouragement throughout the development of this work.
\section{References}
\renewcommand\refname{}


\begin{thebibliography}{99}

\bibitem{evci2021sparsity}
T.~Hoefler, D.~Alistarh,T.~Ben-Nun,N.~Dryden and A.~Peste, \emph{Sparsity in Deep Learning: Pruning and Growth for Efficient Inference and Training in Neural Networks}, arXiv preprint arXiv:2102.00554, 2021. [Online]. Available: \url{https://arxiv.org/abs/2102.00554}

\bibitem{dettmers2019brain}
Z.~Atashgahi,J.~Pieterse, S.~Liu,D.~Mocanu,R.~Veldhuis,and M.~Pechenizkiy, \emph{A Brain-Inspired Algorithm for Training Highly Sparse Neural Networks}, arXiv preprint arXiv:1903.07138, 2019. [Online]. Available: \url{https://arxiv.org/abs/1903.07138}

\bibitem{kirsch2018modular}
L.~Kirsch, J.~Kunze, and D.~Barber, \emph{Modular Networks: Learning to Decompose Neural Computation}, arXiv preprint arXiv:1811.05249, 2018. [Online]. Available: \url{https://arxiv.org/abs/1811.05249}

\bibitem{mcmahan2017communication}
H.~McMahan, E.~Moore, D.~Ramage, S.~Hampson, and B.~Arcas, \emph{Communication-efficient learning of deep networks from decentralized data}, in \emph{Proceedings of the 20th International Conference on Artificial Intelligence and Statistics}, PMLR, 2017, pp. 1273--1282.

\bibitem{li2018federated}
T.~Li, A.~T.~Sahu, M.~Zaheer, M.~Sanjabi, V.~Smith, and A.~Talwalkar, \emph{Federated learning: Challenges, methods, and future directions}, IEEE Signal Processing Magazine, vol. 35, no. 4, pp. 50--60, 2018.
[Online]. Available: \url{https://arxiv.org/pdf/1908.07873}

\bibitem{bohm2018neuron}
E.~Bengio, 
\emph{Reinforcement Learning for Deep Neural Architectures: Conditional Computation with Stochastic Computation Policies}. PhD thesis. McGill University Libraries, 2017.
\bibitem{franklin2019brain}
E.~Bengio, P.~Bacon, J.~Pineau and D.~Precup,
\emph{Conditional computation in neural networks for faster models} in Proc. ICLR, 2016.

\bibitem{liu2021modular}
W.~Fedus, B.~Zoph and N.~Shazeer,
\emph{Switch Transformers: Scaling to Trillion Parameter Models with Simple and Efficient Sparsity," in Journal of Machine Learning Research}, vol. 23, 2022. [Online]. Available: \url{(https://arxiv.org/pdf/2101.03961)}

\bibitem{gallant2019neurosymbolic}
S.~Gross, M.~Ranzato and A.~Szlam, 
\emph{Hard Mixtures of Experts for Large Scale Weakly Supervised Vision} in Proceedings CVPR, 2017.
\end{thebibliography}
\end{document}